\documentclass[letterpaper, 10 pt, conference]{IEEEtran}
\IEEEoverridecommandlockouts                          
\usepackage{graphics} 
\usepackage{epsfig}
\usepackage{times}
\usepackage{amsmath}
\usepackage{amssymb}
\usepackage{threeparttable}
\usepackage{booktabs}
\usepackage{soul}
\usepackage{subfigure}
\usepackage{xcolor}
\usepackage{algorithm}
\usepackage{algorithmic,comment}
\usepackage{tcolorbox}
\usepackage{multirow}
\usepackage{multicol}
\usepackage{cite}
\usepackage[hidelinks]{hyperref}

\begin{document}

\title{\LARGE \bf
TATIC: \underline{T}ask-\underline{A}ware \underline{T}emporal Learning for Human \underline{I}ntent Inference from Physical \underline{C}orrections in Human-Robot Collaboration
}

\author{Jiurun Song$^{1}$, Xiao Liang$^{2}$, and Minghui Zheng$^{3}$%
\thanks{*This work was supported by the USA National Science Foundation under Grant No. 2422826 and 2527316.}%
\thanks{$^{1}$Department of Mechanical Engineering, Texas A\&M University, College Station, TX 77843, USA {\tt\small jiurun\_song@tamu.edu}}%
\thanks{$^{2}$Department of Civil and Environmental Engineering, Texas A\&M University, College Station, TX 77843, USA {\tt\small xliang@tamu.edu}}%
\thanks{$^{3}$Department of Mechanical Engineering, Texas A\&M University, College Station, TX 77843, USA {\tt\small mhzheng@tamu.edu}}}

\maketitle
\thispagestyle{empty}
\pagestyle{empty}


\begin{abstract}
In human–robot collaboration (HRC), robots must adapt online to dynamic task constraints and evolving human intent. While physical corrections provide a natural, low-latency channel for operators to convey motion-level adjustments, extracting task-level semantic intent from such brief interactions remains challenging. Existing foundation-model-based approaches primarily rely on vision and language inputs and lack mechanisms to interpret physical feedback. Meanwhile, traditional physical human–robot interaction (pHRI) methods leverage physical corrections for trajectory guidance but struggle to infer task-level semantics. To bridge this gap, we propose TATIC, a unified framework that utilizes torque-based contact force estimation and a task-aware Temporal Convolutional Network (TCN) to jointly infer discrete task-level intent and estimate continuous motion-level parameters from brief physical corrections. Task-aligned feature canonicalization ensures robust generalization across diverse layouts, while an intent-driven adaptation scheme translates inferred human intent into robot motion adaptations. Experiments achieve a 0.904 Macro-F1 score in intent recognition and demonstrate successful hardware validation in collaborative disassembly (see \textcolor{blue}{\href{https://youtu.be/xF8A52qwEc8}{experimental video}}).
\end{abstract}


\section{Introduction}

\begin{figure}[!t]
    \centering
    \includegraphics[width=0.9\linewidth]{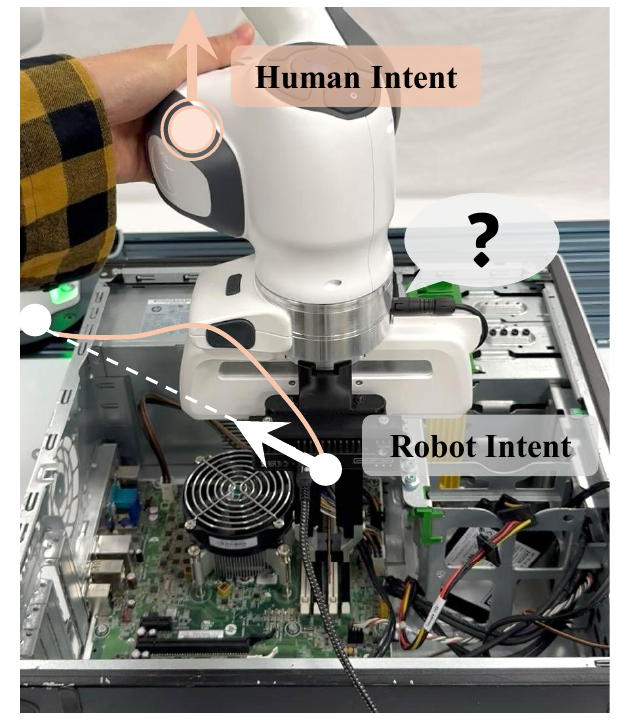}
    \caption{Collaborative desktop disassembly example. During RAM removal, limited visibility may prevent the robot from anticipating collisions with surrounding components. A brief physical correction conveys human intent and triggers motion adaptation to avoid the obstruction.}
    \label{fig:Introduction}
\end{figure}

Robots are increasingly deployed in shared workspaces beyond structured industrial environments, supporting manufacturing, disassembly, and healthcare assistance \cite{arents2022smart, liu2026raise, kyrarini2021survey}. Conventional motion planners typically execute trajectories optimized under predefined objectives and constraints. However, in HRC, such predefined plans often become inadequate since task requirements may change during execution due to evolving human preferences and environmental uncertainty, as illustrated in Fig.~\ref{fig:Introduction}. Therefore, safe and efficient interaction requires robots to accurately infer human intent and adapt their behavior online, thereby enabling responsive collaboration \cite{hoffman2024inferring, losey2022physical, song2025adaptive}.

pHRI establishes a direct communication channel between humans and robots. When a robot deviates from the human's latent objective, the operator can push or pull the robot to deform its trajectory \cite{bajcsy2017learning, losey2017trajectory}. Most existing approaches model these physical corrections as communication signals, using them to update reward functions or objective parameters online \cite{li2021learning, korkmaz2025mile, bobu2020quantifying}. However, relying on such physical feedback often requires continuous kinesthetic guidance, which can induce physical and cognitive fatigue during long-horizon tasks. In addition, while physical interactions naturally encode information related to low-level motion adjustments, extracting higher-level semantic intent from brief contact remains challenging.

While pHRI methods provide limited semantic abstraction, recent progress in semantic planning and robot control has been driven by large language models (LLMs) and vision-language models (VLMs), leading to end-to-end Vision--Language--Action (VLA) policies \cite{driess2023palm, team2024octo, brohan2022rt, zitkovich2023rt}. Trained on large-scale datasets, VLA policies have demonstrated broad generalization, mapping visual observations and natural language instructions directly to robot actions \cite{kim2024openvla, black2024pi0}. However, current VLA architectures primarily operate on vision and language observations, without explicitly modeling force feedback \cite{yu2025forcevla}. In HRC scenarios such as collaborative assembly and disassembly, visual occlusions and high-precision dexterous manipulation requirements can render purely vision–language policies inadequate \cite{liu2025vision}. In these scenarios, brief physical corrections offer a low-latency interaction channel that can convey both task-level intent changes and motion-level adjustments, complementing vision–language inputs \cite{zhang2024don, wang2025inference, xu2025compliant}.

The preceding discussion highlights a gap in existing approaches for HRC: foundation models focus on high-level semantic reasoning, while physical interaction methods primarily address low-level motion adaptation. The dual role of physical contact as both a geometric cue for motion modification and a signal for task-level intent remains underexplored. To bridge this gap, this paper introduces \textbf{TATIC}, a unified framework that jointly decodes semantic intent and motion-level corrections from brief physical interactions. By leveraging torque-based contact estimation, TATIC supports brief physical corrections rather than sustained kinesthetic guidance, eliminating external force sensors and reducing operator fatigue. The main contributions are as follows:
\begin{itemize}
\item \textbf{A task-aware temporal learning framework} that uses torque-based contact force estimation and a causal multi-task TCN to jointly infer discrete task-level intent and continuous motion-level geometric parameters from brief physical corrections.
\item \textbf{A task-aligned feature canonicalization method} that projects interaction data into a canonical local frame, decoupling features from global spatial dependencies and thereby ensuring generalization across diverse workspace layouts.
\item \textbf{An intent-driven motion adaptation scheme} that translates the inferred intent and geometric parameters into robot motion primitives, bridging high-level semantic planning and low-level robot execution in HRC.
\end{itemize}


\section{Related Work}

\subsection{Semantic Planning and Control using Foundation Models}

Foundation-model-based approaches have advanced semantic planning and control in robotics. Early approaches used LLMs as high-level task orchestrators, composing predefined skills into long-horizon plans \cite{liang2023codeasp, ahn2022can}. More recent VLA policies learn end-to-end mappings from visual observations and natural language instructions to robot actions \cite{brohan2022rt, zitkovich2023rt, kim2024openvla,black2024pi0}. Large-scale cross-embodiment pretraining improves generalization across objects, scenes, and robot platforms \cite{o2024open, team2024octo, bousmalis2023robocat}. Despite these advances, current VLA architectures remain predominantly exteroceptive, relying mainly on vision and language observations while rarely incorporating force feedback as an explicit modality \cite{yu2025forcevla}. Consequently, physical interaction signals are often handled through 
low-level compliance or force-control mechanisms rather than being explicitly exploited for task-level corrective guidance.


\subsection{Intent Inference from Physical Corrections}

Modern pHRI approaches treat physical corrections as a communication channel for revealing human latent objectives rather than as external disturbances \cite{albu2003cartesian, losey2022physical, bajcsy2017learning}. Physical corrections can be interpreted as signals for trajectory deformation or for updating objective or reward parameters, enabling online motion adaptation during execution \cite{losey2017trajectory, li2021learning}. Recent work further incorporates safety-aware constraints and intent estimation to enable shared autonomy \cite{shao2024constraintaware}. However, existing approaches typically assume sustained kinesthetic corrections, which may increase operator burden. More recent studies integrate physical feedback with semantic planners and generalist policies: Zhang \emph{et al.} use physical corrections to refine language-model-powered task reasoning \cite{zhang2024don}, and Wang \emph{et al.} steer generative policy sampling at inference time through human interactions \cite{wang2025inference}. However, these approaches often treat physical corrections as episodic interventions that trigger task-goal adjustments, rather than explicitly decoding task-level semantic intent and motion-level geometric corrections from physical interactions.

\begin{figure*}[!t]
    \centering
    \includegraphics[width=0.96\linewidth]{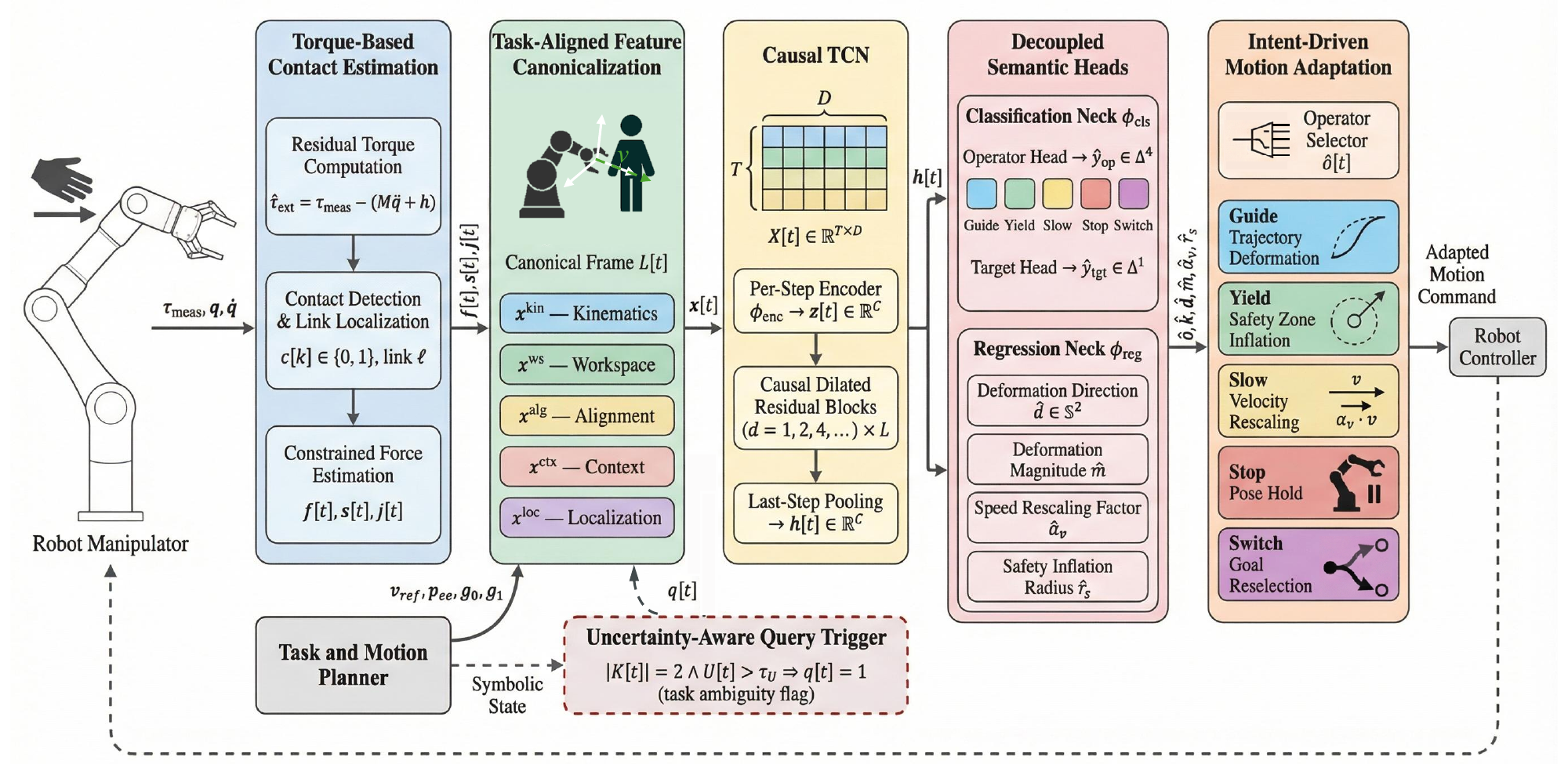}
    \caption{Overview of the proposed TATIC framework.}
    \label{fig:Methodology}
\end{figure*}


\section{Problem Preliminaries}
\label{sec:preliminaries}

\subsection{Robot Dynamics with External Contact}
\label{subsec:robot_dynamics}

Consider an $n$-DoF robot manipulator with joint position $\mathbf{q}\in\mathbb{R}^{n}$, joint velocity $\dot{\mathbf{q}}\in\mathbb{R}^{n}$, joint acceleration $\ddot{\mathbf{q}}\in\mathbb{R}^{n}$, and measured joint torques $\boldsymbol{\tau}_{\mathrm{meas}}\in\mathbb{R}^{n}$. Rigid body dynamics in the presence of external contact satisfy
\begin{equation}
\mathbf{M}(\mathbf{q})\ddot{\mathbf{q}}+\mathbf{h}(\mathbf{q},\dot{\mathbf{q}})+\boldsymbol{\tau}_{\mathrm{ext}}
=
\boldsymbol{\tau}_{\mathrm{meas}},
\label{eq:prelim_dynamics_ext}
\end{equation}
where $\mathbf{M}(\mathbf{q})\in\mathbb{R}^{n\times n}$ is the inertia matrix and $\mathbf{h}(\mathbf{q},\dot{\mathbf{q}})\in\mathbb{R}^{n}$ collects Coriolis and centripetal effects, gravity, and joint friction. The contact model assumes a single contact interaction, represented by an equivalent Cartesian force
$\mathbf{f}_c\in\mathbb{R}^{3}$ applied at a contact location $\mathbf{p}_c\in\mathbb{R}^{3}$. The generalized contact torque follows from virtual work,
\begin{equation}
\boldsymbol{\tau}_{\mathrm{ext}}
=
\mathbf{J}_c(\mathbf{q},\mathbf{p}_c)^\top \mathbf{f}_c,
\label{eq:prelim_tau_ext_jacobian}
\end{equation}
where $\mathbf{J}_c(\mathbf{q},\mathbf{p}_c)\in\mathbb{R}^{3\times n}$ is the translational Jacobian at $\mathbf{p}_c$.


\subsection{Torque-Based Contact Estimation}
\label{subsec:torque_based_force_estimation}

\subsubsection{Residual Torque and Contact Detection}

Let $k\in\mathbb{Z}_{\ge 0}$ denote the sample index. The external torque estimate is
\begin{equation}
\hat{\boldsymbol{\tau}}_{\mathrm{ext}}[k]
=
\boldsymbol{\tau}_{\mathrm{meas}}[k]
-
\left(\mathbf{M}(\mathbf{q}[k])\ddot{\mathbf{q}}[k]+\mathbf{h}(\mathbf{q}[k],\dot{\mathbf{q}}[k])\right),
\label{eq:prelim_residual_torque}
\end{equation}
and the detection statistic
\begin{equation}
\eta[k]=\left\|\mathbf{W}_{\tau}\hat{\boldsymbol{\tau}}_{\mathrm{ext}}[k]\right\|_2,
\label{eq:prelim_residual_stat}
\end{equation}
with diagonal weighting $\mathbf{W}_{\tau}$. An exponentially weighted moving average $\bar{\eta}[k]$ is compared against threshold $\theta_{\tau}$. Contact is declared if $\bar{\eta}[k]>\theta_{\tau}$ for $N_{\mathrm{on}}$ consecutive samples and cleared if $\bar{\eta}[k]<\theta_{\tau}$ for $N_{\mathrm{off}}$ samples. This persistence logic yields a binary contact state $c[k]\in\{0,1\}$.


\subsubsection{Coarse Link Localization}
\label{subsubsec:coarse_link_localization}

Let $\hat{\tau}_{\mathrm{ext},j}[k]$ denote the $j$-th component of $\hat{\boldsymbol{\tau}}_{\mathrm{ext}}[k]$. A joint residual is considered active when $\left|\hat{\tau}_{\mathrm{ext},j}[k]\right|>\tau_{\mathrm{th}}$. The contacted link index is
\begin{equation}
\ell[k]
=
\max
\left\{
j \;\middle|\;
\left|\hat{\tau}_{\mathrm{ext},j}[k]\right|>\tau_{\mathrm{th}}
\land
\left|\hat{\tau}_{\mathrm{ext},j+1}[k]\right|<\tau_{\mathrm{th}}
\right\},
\label{eq:prelim_link_identify}
\end{equation}
with $\ell[k]=n$ if $\left|\hat{\tau}_{\mathrm{ext},n}[k]\right|>\tau_{\mathrm{th}}$. For semantic index $t$, the localized link is denoted $\ell[t]$ and encoded as $\mathbf{j}[t]\in\{0,1\}^{L}$.


\subsubsection{Constrained Force and On-Link Position Estimation}

Conditioned on the contacted link $\ell[t]$, the contact location is parameterized along the link centerline by a normalized scalar $s[t]\in[0,1]$, where $s[t]=0$ corresponds to the link base and $s[t]=1$ corresponds to the link tip. Over a short estimation window $\{k_r\}_{r=1}^{N}$, the contact force is approximated as piecewise constant and denoted by $\mathbf{f}[t]$, allowing the force to vary across different windows within a contact episode. The residual at sample $k_r$ is
\begin{equation}
\mathbf{r}_r(s,\mathbf{f})
=
\hat{\boldsymbol{\tau}}_{\mathrm{ext}}[k_r]
-
\mathbf{J}_c(\mathbf{q}[k_r],\mathbf{p}_c(s,k_r))^\top \mathbf{f}.
\label{eq:prelim_residual_def}
\end{equation}
The contact location $s[t]$ and force $\mathbf{f}[t]$ are obtained via
\begin{equation}
\begin{aligned}
\left[s[t],\mathbf{f}[t]\right]
&=
\arg\min_{s,\mathbf{f}}
\frac{1}{2}\sum_{r=1}^{N}
\left\|\mathbf{r}_r(s,\mathbf{f})\right\|_2^2 \\
&\text{s.t.}\quad
0\le s\le 1,\quad
\|\mathbf{f}\|_2\le F_{\max},
\end{aligned}
\label{eq:prelim_force_estimation_opt}
\end{equation}
The problem is nonconvex in $s$ but linear in $\mathbf{f}$. For fixed $s$, $\mathbf{f}$ is estimated via Tikhonov-regularized least squares and projected onto $\|\mathbf{f}\|_2\le F_{\max}$. The resulting one-dimensional cost over $s\in[0,1]$ is evaluated on a uniform grid and refined using Brent search.


\section{Methodology}

\subsection{System Overview}
\label{subsec:system_overview}

Let $\mathbf{x}[t]\in\mathbb{R}^{D}$ denote the feature vector at semantic time $t$, where $D$ is the feature dimension. Given a fixed-length causal window of $T$ steps, the network input is
\begin{equation}
\mathbf{X}[t]
=
\begin{bmatrix}
\mathbf{x}[t-T+1]^\top \\[-2pt]
\vdots \\[-2pt]
\mathbf{x}[t]^\top
\end{bmatrix}
\in \mathbb{R}^{T \times D}.
\label{eq:window}
\end{equation}
A causal TCN parameterized by $\theta$ maps $\mathbf{X}[t]$ to a joint discrete--continuous output tuple
\begin{equation}
\left[\hat{\mathbf{y}}_{\mathrm{op}}[t],\hat{\mathbf{d}}[t],\hat{m}[t],\hat{\alpha}_{v}[t],\hat{r}_{s}[t],\hat{\mathbf{y}}_{\mathrm{tgt}}[t]\right]
=
\mathcal{F}_{\theta}\!\left(\mathbf{X}[t]\right),
\label{eq:model_map}
\end{equation}
where the discrete operator represents a semantic interaction mode conveyed through brief physical corrections. Accordingly, $\hat{\mathbf{y}}_{\mathrm{op}}[t]\in\mathbb{R}^{5}$ denotes the class posterior over the five operator classes and $\hat{\mathbf{y}}_{\mathrm{tgt}}[t]\in\mathbb{R}^{2}$ denotes the class posterior over the two target candidates, as defined in Sec.~\ref{subsec:taxonomy}. The remaining outputs in \eqref{eq:model_map} correspond to operator-conditioned continuous parameters, as defined in Sec.~\ref{subsec:feature_canonicalization}. Semantic inference is performed over contact segments detected by the contact state $c[k]$. A planner-derived query context variable $q[t]\in\{0,1\}$, computed as described in Sec.~\ref{subsec:Uncertainty}, is appended as an input feature. Fig.~\ref{fig:Methodology} illustrates the overall TATIC framework. For clarity, small positive constants are omitted from subsequent equations and are assumed to be added where needed to ensure numerical stability.


\subsection{Semantic Interaction Taxonomy}
\label{subsec:taxonomy}

The operator set is
\begin{equation}
\mathcal{O}=\left\{\textsc{Guide},\textsc{Yield},\textsc{Slow},\textsc{Stop},\textsc{Switch}\right\}.
\label{eq:operator_set}
\end{equation}
Let $o[t]\in\mathcal{O}$ denote the operator active during the window ending at $t$. Operator semantics are parameterized by operator-conditioned targets with explicit domains. For $o[t]=\textsc{Guide}$, the semantic parameters are a unit direction $\mathbf{d}[t]\in\mathbb{S}^2$ and a normalized displacement magnitude $m[t]\in[0,1]$. For $o[t]=\textsc{Yield}$, the semantic parameter is a normalized safety-margin variable that is mapped to a physical inflation radius at deployment. For $o[t]=\textsc{Slow}$, the nominal execution speed is scaled by $\alpha_{v}[t]\in[\alpha_{\min},\alpha_{\max}]$. For $o[t]=\textsc{Switch}$, the target selection variable satisfies $k[t]\in\{0,1\}$. No auxiliary regression target is defined for $o[t]=\textsc{Stop}$. The discrete operator is decoded by maximum a posteriori (MAP) selection from $\hat{\mathbf{y}}_{\mathrm{op}}[t]$. For \textsc{Switch}, the target index is decoded analogously from $\hat{\mathbf{y}}_{\mathrm{tgt}}[t]$.


\subsection{Feature Canonicalization}
\label{subsec:feature_canonicalization}

Feature construction targets robustness to global translation and yaw while preserving task-relevant interaction geometry. The canonical frame is aligned with the reference motion direction so that semantically equivalent contacts map to similar feature trajectories under changes in end effector pose.

Let $\mathbf{p}_{\mathrm{ee}}[t]\in\mathbb{R}^{3}$ and $R_{\mathrm{ee}}[t]\in SO(3)$ denote end effector position and orientation in world frame $\mathcal{W}$. Let $\mathbf{v}_{\mathrm{ref}}[t]\in\mathbb{R}^{3}$ denote the planner-provided reference velocity and let $\mathbf{f}[t]\in\mathbb{R}^{3}$ denote the estimated contact force from \eqref{eq:prelim_force_estimation_opt}. A local canonical frame $\mathcal{L}[t]$ is constructed from the direction of $\mathbf{v}_{\mathrm{ref}}[t]$ and the world vertical axis $\hat{\mathbf{z}}\in\mathbb{S}^{2}$. The forward axis is initialized by the first nonzero reference velocity direction and is updated with a low-speed fallback,
\begin{equation}
\mathbf{e}_{1}[t]=
\begin{cases}
\dfrac{\mathbf{v}_{\mathrm{ref}}[t]}{\left\|\mathbf{v}_{\mathrm{ref}}[t]\right\|_2}, & \left\|\mathbf{v}_{\mathrm{ref}}[t]\right\|_2>\varepsilon_{\mathrm{ref}},\\[6pt]
\mathbf{e}_{1}[t-1], & \left\|\mathbf{v}_{\mathrm{ref}}[t]\right\|_2\le\varepsilon_{\mathrm{ref}},
\end{cases}
\qquad t\ge 1,
\label{eq:local_forward_fallback}
\end{equation}
where $\varepsilon_{\mathrm{ref}}>0$. To avoid degeneracy when $\mathbf{e}_{1}[t]$ is nearly parallel to $\hat{\mathbf{z}}$, an auxiliary axis is selected as
\begin{equation}
\mathbf{a}[t]=
\begin{cases}
\hat{\mathbf{z}}, & \left\|\hat{\mathbf{z}}\times \mathbf{e}_{1}[t]\right\|_2>\varepsilon_{\times},\\
R_{\mathrm{ee}}[t]\mathbf{e}_y, & \left\|\hat{\mathbf{z}}\times \mathbf{e}_{1}[t]\right\|_2\le\varepsilon_{\times},
\end{cases}
\label{eq:local_aux_axis}
\end{equation}
where $\varepsilon_{\times}>0$ and $\mathbf{e}_y=\left[0,1,0\right]^\top$. The raw orthonormal basis is
\begin{equation}
\mathbf{e}_{2}[t]=\frac{\mathbf{a}[t]\times \mathbf{e}_{1}[t]}{\left\|\mathbf{a}[t]\times \mathbf{e}_{1}[t]\right\|_2},
\qquad
\mathbf{e}_{3}[t]=\mathbf{e}_{1}[t]\times \mathbf{e}_{2}[t],
\label{eq:local_basis}
\end{equation}
and $\tilde{E}[t]=\left[\mathbf{e}_{1}[t]\ \mathbf{e}_{2}[t]\ \mathbf{e}_{3}[t]\right]\in\mathbb{R}^{3\times 3}$. If $\left\|\mathbf{a}[t]\times \mathbf{e}_{1}[t]\right\|_2$ is below a small numerical tolerance, the update is skipped and $\tilde{E}[t]$ is set to $E[t-1]$.

To suppress discontinuities after low speed intervals, the deployed frame $E[t]\in SO(3)$ is smoothed on $SO(3)$ by geodesic interpolation,
\begin{equation}
E[t]
=
E[t-1]\exp\!\left(\lambda_E\,\log\!\left(E[t-1]^\top \tilde{E}[t]\right)\right),
\label{eq:local_frame_smoothing}
\end{equation}
with $E[0]=\tilde{E}[0]$ and $\lambda_E\in(0,1]$. In \eqref{eq:local_frame_smoothing}, $\log(\cdot)$ and $\exp(\cdot)$ denote the Lie logarithm and exponential maps between $SO(3)$ and $\mathfrak{so}(3)$. Projection into $\mathcal{L}[t]$ is defined by $\mathrm{proj}_{t}(\mathbf{v})=E[t]^\top\mathbf{v}$.

Interaction quantities are expressed in the canonical frame $\mathcal{L}[t]$ to remove global pose dependence while preserving task-relevant geometry. The kinematic block includes the reference speed magnitude 
$v_{\mathrm{mag}}[t]=\|\mathbf{v}_{\mathrm{ref}}[t]\|_2$, 
the local unit force direction 
$\hat{\mathbf{f}}_{\mathrm{loc}}[t]$ obtained by normalizing $\mathrm{proj}_{t}(\mathbf{f}[t])$, 
and the force magnitude $\|\mathbf{f}[t]\|_2$. A hemispherical operation space centered at $\mathbf{O}_{hw}\in\mathbb{R}^{3}$ with nominal radius $R_{hw}>0$ and pole direction $\hat{\mathbf{n}}_{hw}\in\mathbb{S}^{2}$ encodes task-level safety constraints independent of instantaneous pose. It consists of points within distance $R_{hw}$ from $\mathbf{O}_{hw}$ whose projection onto $\hat{\mathbf{n}}_{hw}$ is nonnegative. Using the estimated contact point at the terminal sample of the semantic window, the workspace distance feature is defined as the radial clearance
\begin{equation}
d_{hw}[t]
=
R_{hw}
-
\left\|\mathbf{p}_c\!\left(s[t],k_t\right)-\mathbf{O}_{hw}\right\|_2.
\label{eq:dhw}
\end{equation}
The associated escape direction is
\begin{equation}
\mathbf{u}_{\mathrm{esc}}[t]
=
\frac{\mathbf{p}_c(s[t],k_t)-\mathbf{O}_{hw}}{\left\|\mathbf{p}_c(s[t],k_t)-\mathbf{O}_{hw}\right\|_2}.
\label{eq:escape_dir}
\end{equation}
The escape alignment score $a_{\mathrm{esc}}[t]$ is computed as the cosine similarity between $\hat{\mathbf{f}}_{\mathrm{loc}}[t]$ and the normalized projection of $\mathbf{u}_{\mathrm{esc}}[t]$ into $\mathcal{L}[t]$.

Additional geometric cues are expressed as alignment scalars. Candidate goal alignment scores $a_{0}[t]$ and $a_{1}[t]$ measure alignment between $\hat{\mathbf{f}}_{\mathrm{loc}}[t]$ and the projected goal direction rays from $\mathbf{p}_{\mathrm{ee}}[t]$ to the candidate goals $\mathbf{g}_{0}[t]$ and $\mathbf{g}_{1}[t]$. The competition feature is $\Delta a[t]=a_{0}[t]-a_{1}[t]$, which captures directional preference relevant to \textsc{Switch}. Contact localization is encoded by the one-hot link indicator $\mathbf{j}[t]\in\{0,1\}^{L}$ and the normalized surface parameter $s[t]\in[0,1]$. The query context variable $q[t]$ is appended as a binary context feature.

The feature vector $\mathbf{x}[t]\in\mathbb{R}^{D}$ concatenates five functional blocks,
\begin{equation}
\mathbf{x}[t]
=
\operatorname{col}
\left(
\mathbf{x}^{\mathrm{kin}}[t],\,
\mathbf{x}^{\mathrm{ws}}[t],\,
\mathbf{x}^{\mathrm{alg}}[t],\,
\mathbf{x}^{\mathrm{ctx}}[t],\,
\mathbf{x}^{\mathrm{loc}}[t]
\right).
\label{eq:feature_concat}
\end{equation}
The kinematic block $\mathbf{x}^{\mathrm{kin}}[t]$ consists of $v_{\mathrm{mag}}[t]$, $\hat{\mathbf{f}}_{\mathrm{loc}}[t]$, and $\|\mathbf{f}[t]\|_2$. The workspace block $\mathbf{x}^{\mathrm{ws}}[t]$ contains $d_{hw}[t]$. The alignment block $\mathbf{x}^{\mathrm{alg}}[t]$ contains $a_{\mathrm{esc}}[t]$, $a_{0}[t]$, $a_{1}[t]$, and $\Delta a[t]$. The context block $\mathbf{x}^{\mathrm{ctx}}[t]$ contains $q[t]$. The localization block $\mathbf{x}^{\mathrm{loc}}[t]$ contains $\mathbf{j}[t]$ and $s[t]$.


\subsection{Human Intent Inference via TCN}
\label{subsec:tcn}

A causal TCN~\cite{bai2018empirical} maps the windowed feature stream $\mathbf{X}[t]\in\mathbb{R}^{T\times D}$ to joint discrete and continuous semantic estimates. Inputs are standardized using training statistics. Independent dropout is applied to the localization subvector $\left[\mathbf{j}[t]^\top\ s[t]\right]^\top$ to mitigate over-reliance on potentially noisy contact localization.

\subsubsection{Feature Embedding}

Let $C$ denote the channel width. A nonlinear encoder $\varphi_{\mathrm{enc}}$ produces per-step embeddings
\begin{equation}
\mathbf{z}[t]=\varphi_{\mathrm{enc}}\!\left(\mathbf{x}[t]\right)\in\mathbb{R}^{C},
\label{eq:frame_encoder}
\end{equation}
forming the embedded window $\mathbf{Z}[t]=[\mathbf{z}[t-T+1],\ldots,\mathbf{z}[t]]$. The encoder is implemented as a compact MLP.

\subsubsection{Temporal Distillation}

Temporal reasoning is performed by $L$ residual blocks with causal dilated convolutions of kernel size $k$ and dilations $\{d_\ell\}_{\ell=1}^{L}$. The dilation schedule ensures that the effective receptive field covers the full window of length $T$. Let $\mathbf{H}^{(L)}[t]\in\mathbb{R}^{T\times C}$ denote the final block output. Last-step pooling yields
\begin{equation}
\mathbf{h}[t]=\left(\mathbf{H}^{(L)}[t]\right)_{T}.
\label{eq:last_step_pool}
\end{equation}

\subsubsection{Decoupled Semantic Heads}

Two nonlinear projections separate classification and regression:
\begin{equation}
\mathbf{h}^{\mathrm{cls}}[t]=\varphi_{\mathrm{cls}}(\mathbf{h}[t]),\qquad
\mathbf{h}^{\mathrm{reg}}[t]=\varphi_{\mathrm{reg}}(\mathbf{h}[t]).
\label{eq:dual_neck}
\end{equation}

Classification heads map $\mathbf{h}^{\mathrm{cls}}[t]$ to logits $\boldsymbol{\ell}[t]\in\mathbb{R}^{5}$ and $\boldsymbol{\ell}^{\mathrm{tgt}}[t]\in\mathbb{R}^{2}$, producing posteriors $\hat{\mathbf{y}}_{\mathrm{op}}[t]\in\Delta^{4}$ and $\hat{\mathbf{y}}_{\mathrm{tgt}}[t]\in\Delta^{1}$ via softmax. The regression head outputs $\tilde{\mathbf{d}}[t]\in\mathbb{R}^{3}$, normalized to $\hat{\mathbf{d}}[t]\in\mathbb{S}^{2}$, together with $\hat{m}[t]$, $\hat{\alpha}_{v}[t]$, and $\hat{r}_{s}[t]$, which are clamped to the admissible domains in Sec.~\ref{subsec:taxonomy}.

\subsubsection{Multi-Objective Optimization}

Training uses masked multi-task losses. Validity masks activate operator-conditioned targets: $\delta^{\mathrm{dir}}$ and $\delta^{\mathrm{disp}}$ for \textsc{Guide}, $\delta^{\mathrm{spd}}$ for \textsc{Slow}, $\delta^{\mathrm{saf}}$ for \textsc{Yield}, and $\delta^{\mathrm{tgt}}$ for \textsc{Switch}. Operator classification uses focal cross-entropy, and target selection uses cross-entropy. Direction regression adopts the masked cosine loss
\begin{equation}
\mathcal{L}_{\mathrm{dir}}[t]
=
\delta^{\mathrm{dir}}[t]
\left(
1-\frac{\hat{\mathbf{d}}[t]^\top \mathbf{d}_{\mathrm{loc}}[t]}
{\|\hat{\mathbf{d}}[t]\|_2\,\|\mathbf{d}_{\mathrm{loc}}[t]\|_2}
\right),
\label{eq:dir_loss}
\end{equation}
with $\mathbf{d}_{\mathrm{loc}}[t]\in\mathbb{S}^{2}$ denoting the ground-truth unit direction in the canonical frame. Scalar regressions use masked mean squared error, while target selection uses masked cross-entropy applied only to \textsc{Switch}.
Homoscedastic uncertainty weighting \cite{kendall2018multi} is used to automatically balance heterogeneous task losses. Let $\mathcal{T}=\{\mathrm{op},\mathrm{dir},\mathrm{disp},\mathrm{spd},\mathrm{saf},\mathrm{tgt}\}$ denote the set of tasks, and let $\bar{\mathcal{L}}_u$ denote the batch-reduced loss for task $u$. Each task is associated with a learnable log-variance parameter $s_u=\log\sigma_u^2$, which implicitly determines its weighting coefficient. The overall objective is
\begin{equation}
\mathcal{L}
=
\sum_{u\in\mathcal{T}}
\left(
e^{-s_u}\,\bar{\mathcal{L}}_{u}+s_u
\right),
\end{equation}
where $\bar{\mathcal L}_u$ corresponds to the task-specific loss. The weighting term $e^{-s_u}$ adaptively scales the contribution of each task according to its estimated uncertainty.


\subsection{Uncertainty-Aware Query Triggering}
\label{subsec:Uncertainty}

Task ambiguity is determined by a deterministic monitor that checks whether the next task target is uniquely defined given the current symbolic state and execution history. The resulting binary indicator $q[t]\in\{0,1\}$ is appended to the feature vector as a task-context variable.

Let $z[t]$ denote the symbolic task state and $\mathcal{H}[t]$ the executed action history. At each evaluation step, the planner provides two candidate next targets indexed by $i\in\{0,1\}$. Candidate admissibility is encoded by $\chi_i(z[t],\mathcal{H}[t])\in\{0,1\}$, and the admissible set is
\begin{equation}
\mathcal{K}[t]
=
\left\{ i\in\{0,1\}\ \middle|\ \chi_i(z[t],\mathcal{H}[t])=1 \right\}.
\label{eq:admissible_targets}
\end{equation}
The cardinality $|\mathcal{K}[t]|$ characterizes planner determinacy: $|\mathcal{K}[t]|=1$ implies a unique next target, $|\mathcal{K}[t]|=2$ indicates ambiguity, and $|\mathcal{K}[t]|=0$ reflects logical inconsistency. When $|\mathcal{K}[t]|=2$, a composite cost $C_i[t]\in\mathbb{R}_{\ge 0}$ is evaluated for each candidate. A planner preference distribution $\boldsymbol{\pi}[t]=[\pi_0[t],\pi_1[t]]^\top$ is formed by applying an admissibility-gated softmin to the composite costs and normalizing the resulting weights so that $\sum_i \pi_i[t]=1$. Ambiguity is quantified by the Shannon entropy of $\boldsymbol{\pi}[t]$ and normalized to $U[t]\in[0,1]$ by the maximum entropy of a binary distribution.
\begin{equation}
U[t]=\frac{H(\boldsymbol{\pi}[t])}{\log 2},\qquad
H(\boldsymbol{\pi})=-\sum_{i=0}^{1}\pi_i\log\pi_i,
\label{eq:planner_ambiguity_score}
\end{equation}
which yields $U[t]\in[0,1]$. The raw query label $\tilde{q}[t]\in\{0,1\}$ is assigned as
\begin{equation}
\tilde{q}[t]=
\begin{cases}
1, & |\mathcal{K}[t]|=0,\\
0, & |\mathcal{K}[t]|=1,\\
\mathbb{I}\!\left(U[t]\ge \tau_U\right), & |\mathcal{K}[t]|=2,
\end{cases}
\label{eq:raw_query_decision}
\end{equation}
with threshold $\tau_U\in[0,1]$. Temporal smoothing and hysteresis are applied to obtain a persistent label $q[t]$.


\subsection{Intent-Driven Motion Adaptation}
\label{subsec:intent_driven_motion_adaptation}

The motion adaptation layer receives the model outputs in \eqref{eq:model_map} and applies an operator conditioned update based on the MAP operator estimate in Sec.~\ref{subsec:taxonomy}. 

\subsubsection{Guide Mode}

The \textsc{Guide} operator applies a bounded deformation to a finite future segment of the nominal end-effector path. Let $\mathbf{x}_d(\sigma)\in\mathbb{R}^{3}$, $\sigma\in[0,1]$, denote the nominal path, and let $\mathbf{d}^{(r)}(\sigma)$ denote the cumulative deformation after update index $r$. The executed path is
\begin{equation}
\tilde{\mathbf{x}}^{(r)}(\sigma)=\mathbf{x}_d(\sigma)+\mathbf{d}^{(r)}(\sigma),
\label{eq:guide_executed_path}
\end{equation}
with $\mathbf{d}^{(0)}(\sigma)\equiv \mathbf{0}$. At update $r$ with terminal semantic index $t_r$, the semantic displacement in world coordinates is
\begin{equation}
\Delta \mathbf{x}^{(r)}_{\mathrm{sem}}
=
m_{\max}\,\bar{m}[t_r]\,E[t_r]\,\hat{\mathbf{d}}[t_r],
\label{eq:guide_semantic_delta}
\end{equation}
where $m_{\max}>0$ bounds the displacement magnitude and $\bar{m}[t_r]\in[0,1]$ is the clipped prediction. Incremental injection avoids drift under sustained contact:
\begin{equation}
\Delta \mathbf{x}^{(r)}_{\mathrm{inc}}
=
\Delta \mathbf{x}^{(r)}_{\mathrm{sem}}-\Delta \mathbf{x}^{(r-1)}_{\mathrm{sem}},
\qquad
\Delta \mathbf{x}^{(0)}_{\mathrm{sem}}=\mathbf{0}.
\label{eq:guide_increment}
\end{equation}
A deadband $\varepsilon\ge 0$ suppresses small updates, yielding $\Delta \mathbf{x}^{(r)}_{\mathrm{db}}$. Let $\sigma_r$ denote the current path parameter. The deformation is restricted to a finite horizon
\begin{equation}
H_r^\star
=
\min\!\left\{\beta\,\|\Delta \mathbf{x}^{(r)}_{\mathrm{sem}}\|_2,\ 1-\sigma_r\right\},
\label{eq:guide_horizon}
\end{equation}
where $\beta>0$ scales the support and truncation by $1-\sigma_r$ preserves the endpoint. For $\sigma\in[\sigma_r,\sigma_r+H_r^\star]$, a normalized coordinate $\xi_r(\sigma)\in[0,1]$ defines a smooth bump function $b(\xi)=64\xi^3(1-\xi)^3$. The incremental deformation is
\begin{equation}
\boldsymbol{\phi}_r(\sigma)
=
\Delta \mathbf{x}^{(r)}_{\mathrm{db}}\,b\!\left(\xi_r(\sigma)\right),
\label{eq:guide_phi}
\end{equation}
with $\boldsymbol{\phi}_r(\sigma)=\mathbf{0}$ outside the support. The cumulative deformation updates as
\[
\mathbf{d}^{(r)}(\sigma)=\mathbf{d}^{(r-1)}(\sigma)+\boldsymbol{\phi}_r(\sigma),
\]
which modifies only a bounded future segment while preserving the nominal path endpoint.

\subsubsection{Yield Mode}

The \textsc{Yield} operator increases a task-centric safety buffer by inflating the nominal operation space. The network outputs a normalized margin $\hat{r}_{s}[t]$, which is converted to a physical inflation radius and clipped to $[r_{\min},r_{\max}]$, yielding $\bar{r}_{s}[t]\in[r_{\min},r_{\max}]$. The inflated safety zone is defined by the Minkowski sum
\begin{equation}
\tilde{\mathcal{S}}_{hw}[t]
=
\mathcal{S}_{hw} \oplus \mathbb{B}\!\left(\bar{r}_{s}[t]\right),
\label{eq:yield_inflation}
\end{equation}
where $\mathbb{B}\!\left(\bar{r}_{s}[t]\right)$ is a closed ball of radius $\bar{r}_{s}[t]$. The inflated obstacle set $\tilde{\mathcal{S}}_{hw}[t]$ is forwarded to the collision-avoidance motion planner.

\subsubsection{Slow Mode}

The \textsc{Slow} operator rescales the nominal reference velocity while preserving nominal path geometry. The commanded reference velocity is
\begin{equation}
\mathbf{v}^{\mathrm{cmd}}_{\mathrm{ref}}[t]
=
\bar{\alpha}_{v}[t]\,\mathbf{v}_{\mathrm{ref}}[t],
\label{eq:slow_mode_scaling}
\end{equation}
where $\bar{\alpha}_{v}[t]\in[\alpha_{\min},\alpha_{\max}]$ is obtained by clipping the predicted speed scaling factor $\hat{\alpha}_{v}[t]$ to deployment bounds $\alpha_{\min}$ and $\alpha_{\max}$. A first order low pass filter is applied to $\bar{\alpha}_{v}[t]$ to ensure compatibility with acceleration limits.

\subsubsection{Switch Mode}

The \textsc{Switch} operator selects a subsequent mission goal from two candidate targets. The selected goal is
\begin{equation}
\mathbf{g}^{\star}[t] = \mathbf{g}_{\hat{k}[t]}[t].
\label{eq:switch_mode_goal}
\end{equation}
The selected goal $\mathbf{g}^{\star}[t]$ is committed as the updated task-level objective. Upon detection of a \textsc{Switch} event, the symbolic planner persists the selected target index, appends the corresponding subtask to the executed history $\mathcal{H}[t]$, and invokes a motion replan.

\subsubsection{Stop Mode}

The \textsc{Stop} operator suspends motion execution and holds the current pose while maintaining an active controller. This software-level hold is distinct from a hardware safety stop, which typically requires rebooting the robot. Instead, execution can resume through simple repeated taps on the robot once the hazardous interaction is resolved.


\begin{table}[t]
\centering
\caption{Per-Class Intent Recognition Performance}
\label{tab:per_class}
\begin{threeparttable}
\begin{tabular}{lccc}
\toprule
Operator & Precision & Recall & F1-Score \\
\midrule
\textsc{Guide}  & $.879{\scriptstyle\pm.014}$ & $.846{\scriptstyle\pm.023}$ & $.862{\scriptstyle\pm.021}$ \\
\textsc{Yield}  & $.894{\scriptstyle\pm.016}$ & $.865{\scriptstyle\pm.020}$ & $.879{\scriptstyle\pm.019}$ \\
\textsc{Slow}   & $.931{\scriptstyle\pm.010}$ & $.921{\scriptstyle\pm.015}$ & $.926{\scriptstyle\pm.013}$ \\
\textsc{Stop}   & $.914{\scriptstyle\pm.012}$ & $.889{\scriptstyle\pm.022}$ & $.901{\scriptstyle\pm.018}$ \\
\textsc{Switch} & $\mathbf{.961}{\scriptstyle\pm.008}$ & $\mathbf{.942}{\scriptstyle\pm.013}$ & $\mathbf{.951}{\scriptstyle\pm.011}$ \\
\midrule
Macro Avg. & $.916{\scriptstyle\pm.007}$ & $.893{\scriptstyle\pm.010}$ & $.904{\scriptstyle\pm.009}$ \\
\bottomrule
\end{tabular}
\end{threeparttable}
\end{table}


\begin{table}[t]
\centering
\caption{Ablation Study on Feature Construction}
\label{tab:feature_ablation}
\setlength{\tabcolsep}{3.5pt} 
\begin{threeparttable}
\begin{tabular}{lcccc}
\toprule
\multirow{2}{*}{Feature Configuration} & Op. F1 & Tgt. F1 & Dir. Cos & Mag. \\
& ($\uparrow$) & ($\uparrow$) & ($\uparrow$) & RMSE ($\downarrow$) \\
\midrule
Kin. only ($\mathbf{x}^{\mathrm{kin}}$) & $.583{\scriptstyle\pm.017}$ & $.524{\scriptstyle\pm.022}$ & $.741{\scriptstyle\pm.019}$ & $.182{\scriptstyle\pm.011}$ \\
$+$ Workspace ($\mathbf{x}^{\mathrm{ws}}$) & $.712{\scriptstyle\pm.014}$ & $.557{\scriptstyle\pm.019}$ & $.803{\scriptstyle\pm.015}$ & $.149{\scriptstyle\pm.009}$ \\
$+$ Alignment ($\mathbf{x}^{\mathrm{alg}}$) & $.841{\scriptstyle\pm.011}$ & $.891{\scriptstyle\pm.013}$ & $.872{\scriptstyle\pm.012}$ & $.112{\scriptstyle\pm.007}$ \\
TATIC (Full) & $\mathbf{.904}{\scriptstyle\pm.009}$ & $\mathbf{.938}{\scriptstyle\pm.010}$ & $\mathbf{.916}{\scriptstyle\pm.008}$ & $\mathbf{.074}{\scriptstyle\pm.005}$ \\
\bottomrule
\end{tabular}
\end{threeparttable}
\end{table}


\begin{table}[t]
\centering
\caption{Ablation Study on Feature Canonicalization}
\label{tab:canonical_ablation}
\begin{threeparttable}
\begin{tabular}{lccc}
\toprule
\multirow{2}{*}{Representation} & Train & ID Test & OOD-Reconfig \\
& F1 ($\uparrow$) & F1 ($\uparrow$) & F1 ($\uparrow$) \\
\midrule
World Frame & $\mathbf{.921}{\scriptstyle\pm.008}$ & $.889{\scriptstyle\pm.011}$ & $.614{\scriptstyle\pm.024}$ \\
World + SE(2) Aug & $.903{\scriptstyle\pm.009}$ & $.891{\scriptstyle\pm.010}$ & $.753{\scriptstyle\pm.019}$ \\
Canonical Frame (Ours) & $.912{\scriptstyle\pm.007}$ & $\mathbf{.904}{\scriptstyle\pm.009}$ & $\mathbf{.871}{\scriptstyle\pm.013}$ \\
\midrule
& \multicolumn{3}{c}{Direction Cosine Similarity ($\uparrow$)} \\
\cmidrule(l){2-4}
World Frame & $\mathbf{.918}{\scriptstyle\pm.006}$ & $.901{\scriptstyle\pm.009}$ & $.573{\scriptstyle\pm.028}$ \\
World + SE(2) Aug & $.907{\scriptstyle\pm.007}$ & $.896{\scriptstyle\pm.010}$ & $.706{\scriptstyle\pm.021}$ \\
Canonical Frame (Ours) & $.914{\scriptstyle\pm.006}$ & $\mathbf{.916}{\scriptstyle\pm.008}$ & $\mathbf{.887}{\scriptstyle\pm.011}$ \\
\bottomrule
\end{tabular}
\end{threeparttable}
\end{table}


\begin{table}[t]
\centering
\caption{Regression Performance}
\label{tab:regression}
\begin{threeparttable}
\setlength{\tabcolsep}{2.5pt}
\begin{tabular}{llccccc}
\toprule
\multirow{2}{*}{Operator} & \multirow{2}{*}{Target} & \multicolumn{3}{c}{Non-temporal} & \multicolumn{2}{c}{TATIC (Ours)} \\
\cmidrule(lr){3-5} \cmidrule(l){6-7}
& & Mean & Lin. & MLP & GT-Filt. & E2E \\
\midrule
\textsc{Guide} & Dir.\ $\hat{\mathbf{d}}$ (cos$\uparrow$) & .031 & .482 & .641 & $\mathbf{.916}{\scriptstyle\pm.008}$ & $.891{\scriptstyle\pm.012}$ \\
\textsc{Guide} & Mag.\ $\hat{m}$ (RMSE$\downarrow$) & .213 & .156 & .124 & $\mathbf{.074}{\scriptstyle\pm.005}$ & $.098{\scriptstyle\pm.007}$ \\
\textsc{Slow} & Spd.\ $\hat{\alpha}_v$ (RMSE$\downarrow$) & .184 & .142 & .113 & $\mathbf{.061}{\scriptstyle\pm.004}$ & $.079{\scriptstyle\pm.006}$ \\
\textsc{Yield} & Rad.\ $\hat{r}_s$ (RMSE$\downarrow$) & .197 & .152 & .126 & $\mathbf{.086}{\scriptstyle\pm.006}$ & $.108{\scriptstyle\pm.008}$ \\
\textsc{Switch} & Tgt.\ $\hat{k}$ (F1$\uparrow$) & .500 & .694 & .781 & $\mathbf{.938}{\scriptstyle\pm.010}$ & $.921{\scriptstyle\pm.013}$ \\
\bottomrule
\end{tabular}
\begin{tablenotes}\footnotesize
\item Mean: constant mean predictor; Lin.: linear model; MLP: multi-layer perceptron.
\end{tablenotes}
\end{threeparttable}
\end{table}


\section{Experiments}
\label{sec:experiments}

\subsection{Data Collection and Training}
\label{subsec:data_collection}

A pHRI dataset was collected on a 7-DoF manipulator, comprising 500 in-distribution (ID) episodes across 100 trajectories with diverse workspace layouts and object orientations. For canonicalization evaluation, an additional disjoint out-of-distribution (OOD) set of 250 episodes across 50 reconfigured trajectories was collected. Across all episodes, contact durations range from 0.53 s to 1.98 s, indicating brief corrective interactions. The ID set was split at the trajectory level into train/validation/test (70/15/15), while the OOD set was reserved exclusively for OOD evaluation. The causal TCN uses channel width $C=128$ and $L=6$ residual blocks with kernel size $k=3$ and exponentially increasing dilations. A two-layer MLP embeds input features, followed by separate classification and regression heads. Batch normalization is used during training and frozen at inference. The model is trained with AdamW using a cosine learning-rate schedule and focal loss for operator classification, and has approximately 0.63M trainable parameters.


\subsection{Quantitative Results and Ablations}
\label{subsec:quantitative_ablation}

\begin{figure*}[!t]
    \centering
    \includegraphics[width=0.99\linewidth]{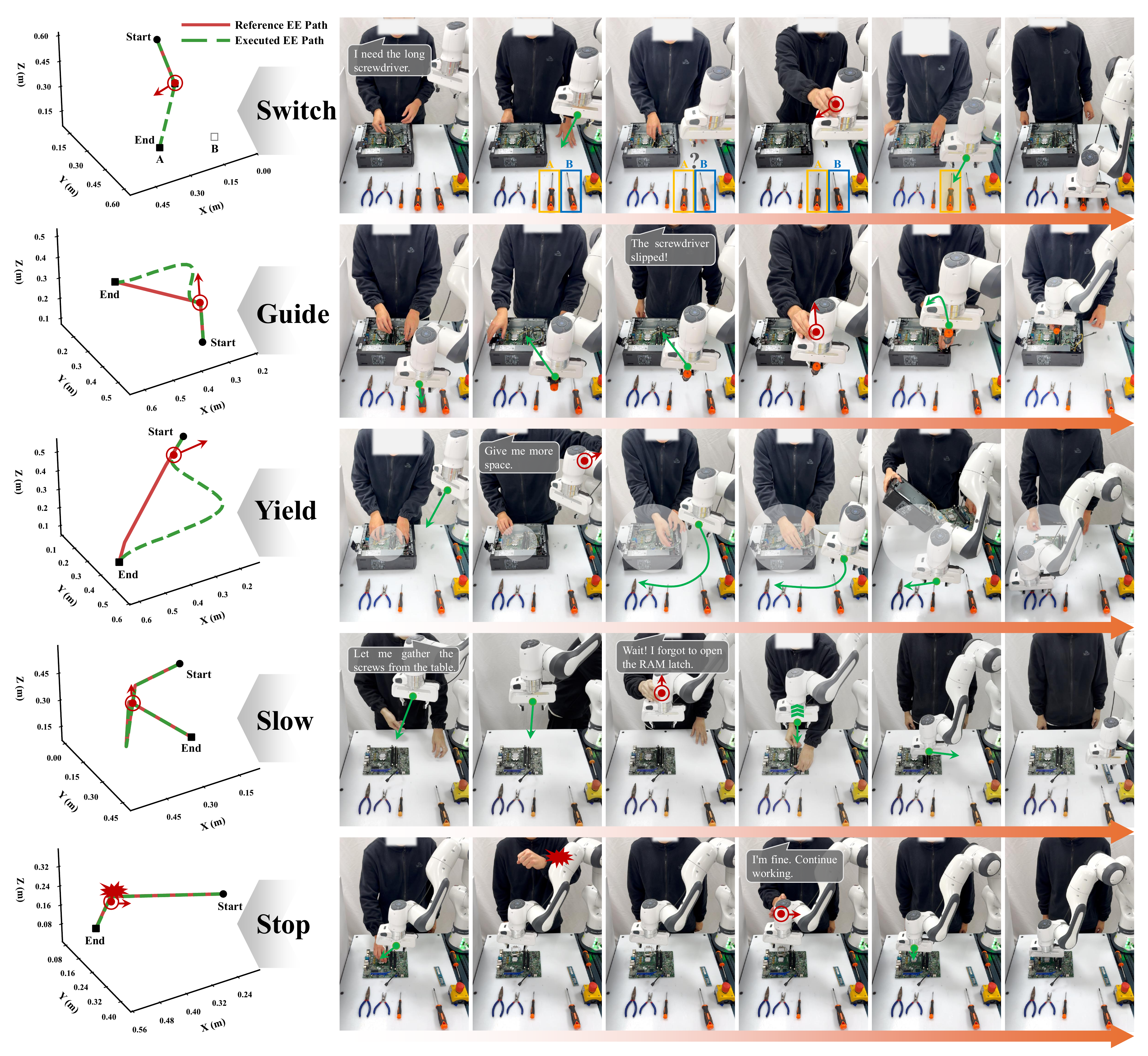}
    \caption{Collaborative desktop disassembly experiment. The human conveys intent through brief physical contact, which is decoded by TATIC into semantic operators and translated into motion adaptation of the robot. Speech-bubble annotations indicate human intent change rather than language prompts. See the supplementary video for the complete execution sequence.}
    \label{fig:Demo}
\end{figure*}

\subsubsection{Per-Class Recognition}

Table~\ref{tab:per_class} reports the test-set performance. 
\textsc{Switch} achieves the highest F1 score of 0.951. 
\textsc{Guide} and \textsc{Yield} are more difficult to distinguish, with F1 scores of 0.862 and 0.879, respectively. 
The difficulty arises because both operators correspond to lateral pushing behaviors, leading to frequent mutual confusions between \textsc{Guide} and \textsc{Yield}. 
The overall Macro-F1 score is 0.904, and the expected calibration error is 0.041. 
Trajectory-level bootstrap analysis with 1,000 resamples yields a 95\% confidence interval of [0.887, 0.922] for the Macro-F1 score.

\subsubsection{Feature Ablation}

Table~\ref{tab:feature_ablation} evaluates the incremental contribution of each feature group. 
Using kinematic features alone yields an operator Macro-F1 of 0.583 and limited separation between \textsc{Guide} and \textsc{Yield}, with a target selection F1 of 0.524. 
Adding workspace cues increases the Macro-F1 to 0.712 and the target F1 to 0.557, indicating improved discrimination of lateral pushing behaviors. This gain may be attributed to the radial clearance feature $d_{hw}$, which encodes boundary proximity absent from kinematics alone.
Incorporating alignment features further boosts performance, raising target F1 to 0.891 and improving \textsc{Guide}-conditioned regression. 
With the full feature set, the model achieves a Macro-F1 of 0.904 and a target F1 of 0.938.

\subsubsection{Canonicalization Ablation}
\label{subsec:canon_ablation}

Table~\ref{tab:canonical_ablation} reports classification F1 and \textsc{Guide}-conditioned direction cosine similarity for three feature representations (world frame, world$+$SE(2) augmentation, and the proposed canonical frame) on the training, ID-test, and OOD-Reconfig splits. 
On OOD-Reconfig, the world-frame baseline drops to an F1 of 0.614, 27.5 percentage points below its ID-test F1 of 0.889. 
SE(2) augmentation improves the OOD-Reconfig F1 to 0.753, and the canonical representation further improves it to 0.871. 
A consistent trend is observed for \textsc{Guide}-conditioned direction cosine similarity. 
Overall, explicit canonicalization better mitigates performance degradation under layout reconfiguration than SE(2) augmentation alone.

\subsubsection{Regression}

Table~\ref{tab:regression} compares regression performance under five settings: three non-temporal baselines (Mean, Lin., and MLP), a GT-Filt. setting with ground-truth operator labels, and an end-to-end (E2E) setting using predicted operators. Using identical input features, the non-temporal baselines underperform the proposed causal TCN across all targets. For \textsc{Guide} direction prediction, E2E achieves a cosine similarity of 0.891, substantially higher than MLP (0.641), while reducing magnitude RMSE from 0.124 to 0.098. Similar gains are observed for \textsc{Slow} and \textsc{Yield}, where E2E RMSE remains lower than all non-temporal baselines. For \textsc{Switch}, target selection improves from 0.781 (MLP) to 0.921 (E2E). The gap between GT-Filt. and E2E reflects operator classification errors, while the strong E2E performance indicates effective temporal reasoning under realistic deployment conditions.


\subsection{Hardware Validation: Collaborative Disassembly}
\label{subsec:case_study}

TATIC is deployed on a 7-DoF manipulator for collaborative desktop disassembly. Target coordinates are provided by a side-mounted and a wrist-mounted RGB-D camera. A constraint-based task planner generates the disassembly sequence from component connectivity and prerequisite relations, and nominal reference motions are generated using cuRobo. The robot performs supportive subtasks (e.g., tool handover and part placement), while the human executes contact-intensive operations such as loosening fasteners and releasing latches. As shown in Fig.~\ref{fig:Demo}, changes in human intent during execution are conveyed through brief physical corrections. Interaction forces are estimated from joint torques, semantic operators and associated parameters are inferred by the temporal model, and motion updates are applied via the adaptation module. Fig.~\ref{fig:Demo} compares nominal and adapted end-effector paths and marks contact onset positions. Overall, reliable semantic inference and responsive motion adaptation are demonstrated in the collaborative disassembly experiment.


\section{Conclusions}

TATIC is proposed as a unified framework for inferring task-level intent and motion-level parameters from brief physical corrections in HRC. Experiments achieve a Macro-F1 of 0.904 for intent recognition and validate closed-loop execution in a collaborative disassembly task. Current limitations include the use of a predefined intent vocabulary and the lack of explicit personalization to individual operators. Future work will investigate adaptive learning to capture user-specific behaviors and to extend the interaction semantics beyond the current predefined set using larger datasets.


\bibliographystyle{IEEEtran} 
\bibliography{ref}        

\end{document}